\documentclass[10pt, a4paper]{article}

\usepackage[final]{lrec2026} 

\usepackage[whole]{bxcjkjatype} 
\usepackage{booktabs} 
\usepackage{multirow} 
\usepackage{amssymb}
\usepackage{enumitem}

\title{JBE-QA: Japanese Bar Exam QA Dataset \\ for Assessing Legal Domain Knowledge}

\name{
Zhihan Cao$^1$, Fumihito Nishino$^2$, Hiroaki Yamada$^1$, \\
{\bf \large Nguyen Ha Thanh$^{2,3}$,  Yusuke Miyao$^4$,  Ken Satoh$^2$}
} 

\address{
$^1$~Institute of Science Tokyo, Japan \quad $^2$~ROIS-DS, Center for Juris-informatics, Japan\\
$^3$~National Institute of Informatics, Japan \quad $^4$~University of Tokyo, Japan\\
cao.z.c8a7@m.isct.ac.jp \quad nishino@nii.ac.jp,\quad yamada@comp.isct.ac.jp \\
nguyenhathanh@nii.ac.jp \quad yusuke@is.s.u-tokyo.ac.jp \quad ksatoh@nii.ac.jp\\
}

\abstract{
We introduce JBE-QA, a Japanese Bar Exam Question–Answering dataset to evaluate large language models' legal knowledge. Derived from the multiple-choice (\textit{tantō-shiki}) section of the Japanese bar exam (2015–2024),  JBE-QA provides the first comprehensive benchmark for Japanese legal-domain evaluation of LLMs. It covers the Civil Code, the Penal Code, and the Constitution, extending beyond the Civil Code focus of prior Japanese resources.
Each question is decomposed into independent true/false judgments with structured contextual fields. 
The dataset contains 3,464 items with balanced labels. We evaluate 26 LLMs, including proprietary, open-weight, Japanese-specialised, and reasoning models. Our results show that proprietary models with reasoning enabled perform best, and the Constitution questions are generally easier than the Civil Code or the Penal Code questions. 
\\ \newline 
\Keywords{Japanese Bar Exam, Legal QA, Japanese law, LLM evaluation, benchmark, dataset} }

\begin{document}

\maketitleabstract

\section{Introduction}
Legal practice relies heavily on the individual professional's expertise and experience. 
This expert-dependency raises the need for legal information processing, which, for example, helps retrieve legal documents efficiently and analyse their argumentative structures~\cite{nguyen2024attentive,vuong2023sm}.
In this way, legal information processing reduces the expert-dependency in legal practice and improves access to legal information for non-expert users.
Currently, Large Language Models~(LLMs) are widely applied to legal information processing~\cite{nguyen2025llms}.
Given the high-stakes nature of legal information processing, LLMs must have accurate knowledge in the legal domain to be reliable.

Previous researchers have developed resources that evaluate the knowledge quality of LLMs.
LegalBench~\cite{guha2023legalbench} is a multi-task benchmark focused on legal reasoning, covering subjects such as the Constitution, the Penal Code, and the Civil Code of the United States.
MultiEURLEX~\cite{chalkidis2021multieurlex} is a multilingual corpus of EU laws for document topic classification, covering laws of transportation and agriculture.
JEC-QA~\cite{zhong2020jec} is a large-scale question-answering dataset derived from the Chinese National Judicial Examination.
It contains questions regarding diverse subjects such as the Constitution, the Penal Code, and the Civil Code.

Nevertheless, the availability, as well as the scope, of Japanese legal resources are limited.
Only a few publicly available datasets exist, most of which focus on the Civil Code of Japan.
For example, the COLIEE dataset~\cite{goebel2024overview} involves information retrieval and entailment tasks concerning the Civil Code.
\citet{2023GPT-barexam} implemented binary legal questions to evaluate LLM's performance in the Japanese legal domain, but they only released a subset for the Civil Code questions.
The Japanese Tort-case Dataset~\cite{yamada2024japanese} evaluates the ability of models to predict judgments in tort cases from the Civil Code.
In summary, the existing Japanese resources have their limitations in coverage: 
they mainly focus on the Civil Code, whereas few target the Penal Code or the Constitution.

Three major subjects of law, the Civil Code, the Penal Code, and the Constitution, differ in functions and positions.
The Civil Code, as a core of private laws, regulates legal interactions between private individuals.
However, since legal issues also arise between the public and the private, the public laws, exemplified by the Penal Code or the Constitution, are equally essential for understanding the law.
Particularly, the Constitution is important as it is the supreme law governing all the laws.
Given the above, all three major subjects of law must be covered in an evaluation resource to establish a comprehensive evaluation of LLMs' legal knowledge.
Yet, there is no currently available resource that fulfills this requirement, leading to a partial and insufficient understanding of LLMs' legal knowledge.

To fill this gap, we construct a new comprehensive dataset, Japanese Bar Exam QA~(JBE-QA).
JBE-QA dataset targets the Civil Code, the Penal Code, and the Constitution at the same time and hence has a larger scope compared to existing evaluation resources in the Japanese legal domain.
In this way, our JBE-QA dataset enables a comprehensive legal knowledge evaluation.
We establish baseline performance on the JBE-QA dataset by evaluating 26 large language models, including proprietary, open-weight, generic multilingual, Japanese-specialised, and reasoning-oriented models.
This baseline demonstrates the coverage of Japanese legal knowledge in current LLMs and serves as a benchmark for future model improvements.
Our dataset is available \href{https://github.com/hancules/JBE-QA}{online}.

\section{Dataset Construction}
Our dataset is derived from the Japanese Bar Examination (JBE)\footnote{司法試験, \textit{Shihō Shiken}}, specifically from the multiple-choice type questions, \textit{tantō-shiki} test.
All the data are in the Japanese language.

\subsection{Japanese Bar Exam}
\begin{figure*}[t]
    \centering
    \fbox{\includegraphics[width=\linewidth]{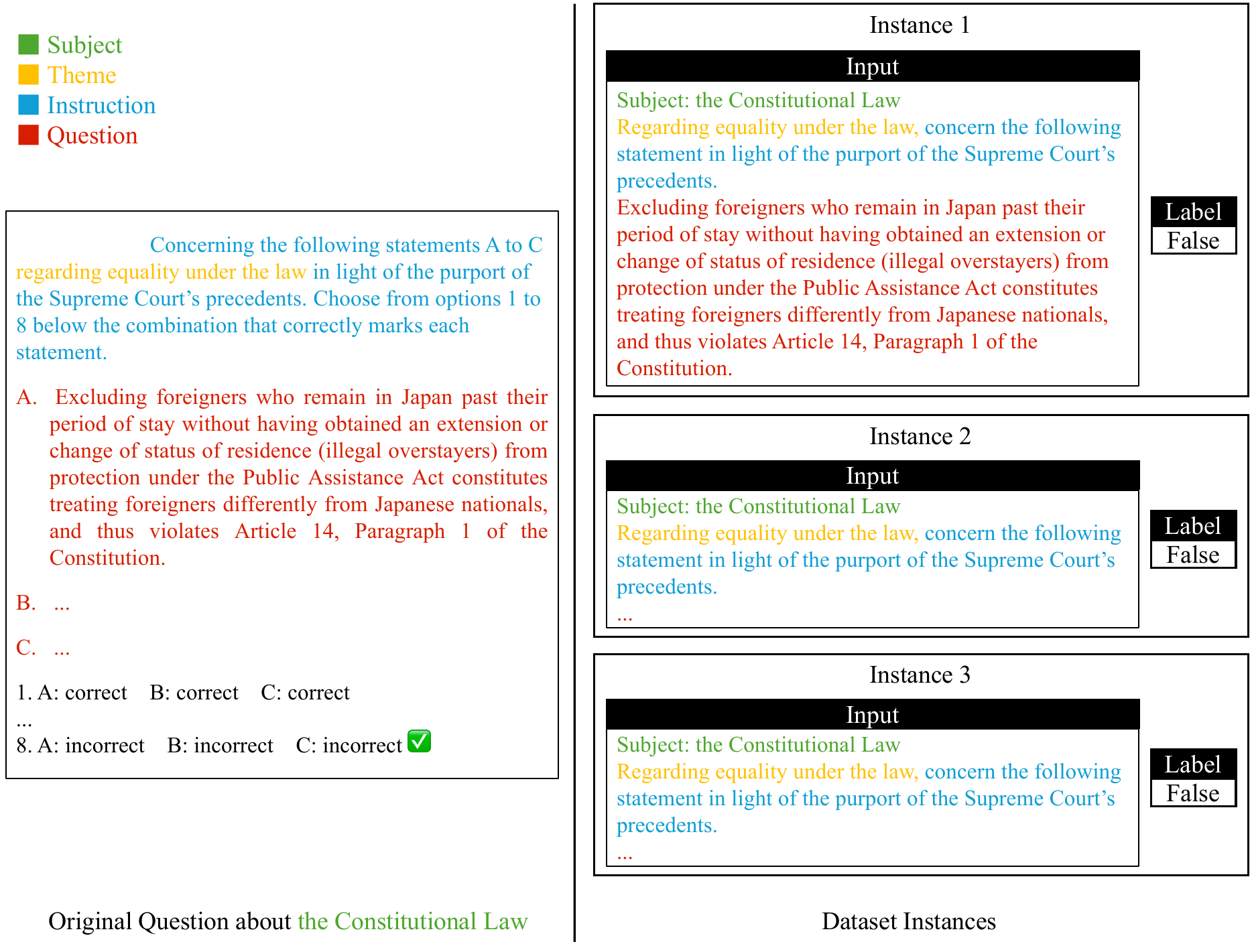}}
    \caption{
    English translation of an example of the Japanese bar exam \textit{tantō-shiki} test (left) and its format in the JBE-QA dataset (right). 
    Note that not all fields introduced in Section~\ref{sec:fields} are in this example.
    }
    \label{fig:example}
\end{figure*}
JBE is a national examination that grants the legal qualifications required to become a judge, prosecutor, or attorney in Japan.
The JBE consists of two tests: the \textit{tantō-shiki} test and the \textit{ronbun-shiki} test, where the former is a multiple-choice test, and the latter is an essay-writing test.
The \textit{tantō-shiki} test assesses whether an examinee has the legal knowledge and reasoning ability required to become a legal professional.
The \textit{ronbun-shiki} test is intended to evaluate whether the examinee has the skills in legal analysis, organisation, and argumentation writing that are necessary for legal practice.
Both tests include a wide range of problems from multiple subjects, including the Civil Code, the Penal Code, and the Constitution.

\subsection{Question Design}
Our JBE-QA dataset is based on the \textit{tantō-shiki} test because it focuses more on legal knowledge itself than on the application of legal knowledge.
This aligns with our evaluation objective of comprehensively assessing LLMs' understanding of legal knowledge.
Furthermore, its question format is choice-based, making it suitable for automated evaluation.
In the original \textit{tantō-shiki} test, a question is usually a set of statements about a legal topic, each of which is either true or false\footnote{
There are minor cases where the question has a more complicated structure.
}.
For a question, multiple options are provided, and each option is a list of binary values that indicate the truth of every statement in the question.
Examinees need to pick the option that correctly represents the truth conditions of all statements in the question.

On the left-hand side of Figure~\ref{fig:example}, we show a question regarding the Constitution. 
Examinees are given multiple-choice options numbered from 1 to 8, each representing a different combination of truth values.
In the example question, as all the statements are incorrect, examinees should select option 8.

In our dataset, we simplify the \textit{tantō-shiki} test by dividing each question into multiple binary classification problems.
A model under evaluation only needs to determine the truth of each statement one at a time, rather than the combinations of truths of a set of statements.
This allows us to interpret the result in a more fine-grained manner and align with the former practice in COLIEE~\citep{goebel2024overview} as well.
In the example, we split the question into three binary classification tasks corresponding to statements A, B, and C, as shown on the right-hand side of Figure~\ref{fig:example}.

\subsection{Data Collection and Preprocessing}
Past exam PDFs are publicly available on the website of the Ministry of Justice of Japan\footnote{\url{https://www.moj.go.jp/barexam.html}}.
We downloaded the PDFs of the past exams between 2015 and 2024.
The PDFs were converted into structured XML files for preprocessing. 
Automated scripts were applied to normalise formats and detect common extraction errors, followed by targeted manual review and correction where necessary.

Some of the questions from the original bar exam could not be formatted as binary true/false items.
We removed such questions, resulting in our final dataset containing minor differences from the original exams. 
In total, 52 questions were excluded: 46 from the Penal Code, 5 from the Constitution, and 1 from the Civil Code.

\subsection{Data Structure and Quality Assurance}
\label{sec:fields}
To convert the original multiple-choice questions into binary questions, it is necessary to segment and then structure the question text.

All the original questions have instructional text, which provides explicit guidance on the basis for answering them~(shown in blue fonts in Figure~\ref{fig:example}). 
We extract this part as \texttt{instruction} field.
Another element that always exists in the original question is statement text shown in red fonts in Figure~\ref{fig:example}.
An LLM must determine the truth of the statement text. 
We hold them as \texttt{question} field in our dataset.
\texttt{Theme} field, shown in yellow fonts in Figure~\ref{fig:example}, gives scope or topics of the questions.

Some of the original questions have auxiliary text providing context or additional materials. 
For example, there is a type of question that their \texttt{instruction} refers to a long description of a case presented separately from the instruction. 
We extract such text as \texttt{lead\_in} field in our dataset.
Also, some questions have clarifications or special remarks that make them simpler or solvable. 
An example is the expression ``Ignore damages, and enforcement costs.'' in a question about identifying the correct amount of money to be distributed to each stakeholder of a real estate. 
\texttt{remark} field holds them.

Building on this structure, each instance in the dataset contains the following fields:
\begin{itemize}[leftmargin=*]
  \itemsep0em 
  \item \texttt{id}: Unique identifier for each question.
  \item \texttt{year}: Examination year in Japanese era format.
  \item \texttt{subject}: Legal subject in English.
  \item \texttt{subject\_jp}: Legal subject in Japanese. 
  \item \texttt{instruction}: Clear instructions for answering in Japanese.
  \item \texttt{question}: Question text in Japanese.
  \item \texttt{label}: Original label (Y for True, N for False).
  \item \texttt{answer}: Standardised answer (True/False).
  \item \texttt{theme}: legal topic or theme in Japanese. Optional.
  \item \texttt{lead\_in}: Auxiliary text provided separately from instruction or question. Optional.
  \item \texttt{remark}: Clarifications or supplementary text. Optional.
\end{itemize}

To ensure the quality and integrity of the dataset, we performed a thorough manual review to identify and correct any mismatches between questions and their gold labels that had occurred during the initial automated extraction. 
This post-processing step was vital for improving the reliability of the dataset as a benchmark for evaluating LLMs.

\subsection{Dataset Statistics}
The dataset consists of 3,464 question-answer pairs.
Table~\ref{tab:subject-distribution} shows the distribution of questions across the three core legal subjects.
This distribution reflects the emphasis on the Civil Code in the Japanese Bar Examination. The dataset is balanced in terms of answer distribution, with 52.4\% of the instances having a ``False'' label and 47.6\% having a ``True'' label. 
\begin{table}[t]
\centering
\small
\begin{tabular}{lcc}
\hline
\textbf{Subject} & \textbf{Count} & \textbf{Percentage} \\
\hline
The Civil Code & 1,998 & 57.7\% \\
The Penal Code & 811 & 23.4\% \\
The Constitution & 655 & 18.9\% \\
\hline
\end{tabular}
\caption{Distribution of questions subjects}
\label{tab:subject-distribution}
\end{table}

\section{Baseline Experiment}
\label{sec:baseline_experiment}
Each instance in the JBE-QA dataset represents a binary classification problem.
We instruct an LLM under evaluation to only output a binary truth value as the prediction for every instance~(c.f. Section~\ref{sec:system_prompt}).
However, some models may fail to follow the instruction, resulting in non-binary outputs.
In such cases, the prediction is set to 0 (False) by default.
In order to quantify the frequency of such cases, we compute the faithfulness to instruction~(hereafter, faithfulness score), defined as the ratio of instances where the model produces a binary output.

We use the F1 score as the metric of performance, which is standard for binary classification.
Each model is evaluated in zero-shot and four-shot settings.
In order to ensure the comparability of results, we exclude instances used as exemplars for the four-shot setting while calculating the results.
Consequently, each model's score under a given metric is averaged over $3464-4=3460$ instances.

\subsection{Models}
We evaluate various types of models, including both proprietary and open-weight ones.
For proprietary models, we consider models from OpenAI~\cite{openai2023gpt4} and Anthropic~\cite{anthropic2024claude3}.
For the OpenAI models, we employ GPT-4.1, GPT-4o, GPT-5, o3, and o4-mini\footnote{
\texttt{gpt-4.1-2025-04-14}, \texttt{gpt-4o-2024-11-20}, \texttt{gpt-5-2025-08-07}, \texttt{o3-2025-04-16}, \texttt{o4-mini-2025-04-16} in order.
}.
Among them, GPT-5, o3, and o4-mini are reasoning models.
For the Anthropic models, we use four Claude models:
Opus-3, Opus-4.1, Sonnet-4, Sonnet-4.5, and Haiku-3.5\footnote{
\texttt{claude-3-opus-20240229}, \texttt{claude-opus-4-1-20250805}, \texttt{claude-sonnet-4-20250514}, \texttt{claude-sonnet-4-5-20250929}, \texttt{claude-3-5-haiku-20241022} in order.
}.
Among them, Claude Opus-4.1 and Sonnet-4~\& 4.5 support reasoning (``extended thinking'')~\cite{anthropic-extended-thinking}, and we evaluate these three models both with and without reasoning enabled.

For open-weight models, we consider both generic multilingual and Japanese-specialised models in order to investigate whether general-domain adaptation for Japanese leads to better performance in the Japanese legal domain as well. 
All the models selected are instruction-tuned and have more than 15 billion parameters. 
For generic models, we employ Qwen 2.5-32B~\& 72B~\cite{qwen2025qwen25technicalreport}, GPT-OSS-20B~\& 120B~\cite{openai2025gptoss120bgptoss20bmodel}, Gemma 3-12B~\& 27B~\cite{gemmateam2025gemma3technicalreport}, Llama 3.1~\cite{meta2024llama}~\& 3.3~\cite{meta2024llama3_3}\footnote{
\texttt{Qwen2.5-32B-Instruct},
\texttt{Qwen2.5-72B-Instruct},
\texttt{gpt-oss-20b},
\texttt{gpt-oss-120b},
\texttt{gemma-3-12b-it},
\texttt{gemma-3-27b-it},
\texttt{Llama-3.1-70B-Instruct},
\texttt{Llama-3.3-70B-Instruct}
in order.
}.

The Japanese-specialised models include
ABEJA-V2~\cite{abeja2025qwen25}, 
Swallow-3.1 \& 3.3~\cite{Fujii:COLM2024, Okazaki:COLM2024, ma2025building}, 
and two LLM-jp-3.1 models~\cite{nii2025llmjp31}, a 13B dense one and a 104B mixture-of-experts one\footnote{
\texttt{ABEJA-Qwen2.5-32b-Japanese-v0.1},
\texttt{Llama-3.1-Swallow-70B-Instruct-v0.3},
\texttt{Llama-3.3-Swallow-70B-Instruct-v0.4},
\texttt{llm-jp-3.1-13b-instruct4},
\texttt{llm-jp-3.1-8x13b-instruct4}
in order.
}.
ABEJA-V2 is a 32B Qwen 2.5 model fine-tuned on a Japanese corpus.
Swallow-3.1~\& 3.3 are based on Llama-3.1-70B and Llama-3.3-70B, respectively, with continual pre-training on Japanese data.
Unlike the models above that are based on generic multilingual models, the LLM-jp-3.1 series is built upon a Japanese-native LLM.

In total, there are 26 models evaluated, considering Opus-4.1, Sonnet-4, and Sonnet-4.5 with and without reasoning enabled as six different models.

\subsection{Experimental Settings}
\paragraph{System Prompt}
\label{sec:system_prompt}
We instruct models using the following system prompt for each model.
The English translation is given below.

\begin{quote}
以下の法律に関する問題を解答せよ。理由や説明は不要。「正しい」と判断した時に1を、「誤り」と判断したときに0を出力せよ。出力は必ず1または0のいずれかの整数値のみとせよ。\\
(
Answer the following legal question. 
No reason or explanation is required. 
Output 1 if you judge the statement to be correct, and 0 if incorrect. 
The output must be only an integer, either 1 or 0.
)
\end{quote}
For llm-jp-3.1 models, which do not allow modifying the system prompt by default, we add the same instruction at the beginning of the user input.

\paragraph{Four-shot Learning Setting}
Four instances are sampled from the training set and used as exemplars across all model evaluation experiments.
Two are randomly sampled from positive instances, and the other two from negative instances.

\paragraph{Question format}
Using the data fields defined in JBE-QA, we formatted questions according to the following rules\footnote{
The formatter script is available online.
}.
All questions are given in Japanese to the models.

The first line specifies the subject of the question~(the \texttt{subject\_jp} field), such as the green text ``Subject: The Constitution'' in Figure~\ref{fig:example}.

The next line indicates the theme of the question~(given in the \texttt{theme} field), if it exists.
In the right-hand side of Figure~\ref{fig:example}, the yellow text ``Regarding equality under the law'' specifies the subject.

Some questions include background information~(stored in the \texttt{lead\_in} field). If so, such information is attached after the theme line.

A question description~(the \texttt{instruction} field) and the statement~(the \texttt{question} field) are presented in order.
In the example, the \texttt{instruction} field is ``concern the following statement in light of the purport of the Supreme Court's precedents''.

Finally, if the question contains additional annotations or clarifications~(the \texttt{remark} field), they are appended at the end.

\paragraph{Hyper-parameters}
We set the temperature to 0 when evaluating non-reasoning models.
This forces the model to predict the token with the highest likelihood at each time step, making generation deterministic.
Non-reasoning models' maximum output length is set to 1,000 tokens.
Note that we expect the models to output either 1 or 0 if they are faithful to our instruction. 
Thus, 1,000 tokens are long enough for our experiments.
For open reasoning models, we use the temperature of 1, and for proprietary reasoning models, we use their default temperature values\footnote{
They are usually 1. 
Some models, for example GPT-5 and o3, do not support adjustment of temperature parameter.
}.
We limit the maximum output length to one quarter of the allowed maximum.
This is in order to ensure the model under evaluation has enough tokens for sufficient reasoning.
All evaluations are conducted once per model.

\section{Baseline Results}
\begin{table*}[htbp]
    \centering
    \small
    \begin{tabular}{llllcccc}
    \toprule
    \multirow{2}{*}{O} & \multirow{2}{*}{J} & \multirow{2}{*}{R} & \multirow{2}{*}{Model} 
    & \multicolumn{2}{c}{F1} & \multicolumn{2}{c}{Faithfulness} \\
    \cmidrule(rl){5-6}\cmidrule(rl){7-8}
     &  &  &  & Zero-shot & Four-shot & Zero-shot & Four-shot \\
    \midrule
    \multirow{7}{*}{$\times$} & \multirow{7}{*}{$\times$} & \multirow{7}{*}{$\times$} 
    & GPT-4.1 & 0.724 & 0.731 & 0.989 & 0.995 \\
    &&& GPT-4o-11 & 0.708 & 0.731 & 0.997 & 0.995 \\
    &&& Sonnet 4 & 0.694 & 0.738 & 0.895 & 1.000 \\
    &&& Opus 3 & 0.686 & 0.708 & 1.000 & 1.000 \\
    &&& Opus 4.1 & 0.602 & 0.799 & 0.648 & 1.000 \\
    &&& Haiku 3.5 & 0.148 & 0.666 & 0.259 & 1.000 \\
    &&& Sonnet 4.5 & 0.055 & 0.751 & 0.027 & 1.000 \\
    \midrule
    \multirow{6}{*}{$\times$} & \multirow{6}{*}{$\times$} & \multirow{6}{*}{$\checkmark$}
    & Opus 4.1 (w/ Reasoning) & 0.814 & 0.861 & 0.873 & 1.000 \\
    &&& GPT-5 & 0.794 & 0.783 & 1.000 & 1.000 \\
    &&& Sonnet 4 (w/ Reasoning) & 0.780 & 0.776 & 0.992 & 1.000 \\
    &&& o3 & 0.769 & 0.762 & 1.000 & 1.000 \\
    &&& o4-mini & 0.672 & 0.673 & 1.000 & 1.000 \\
    &&& Sonnet 4.5 (w/ Reasoning) & 0.077 & 0.828 & 0.040 & 0.999 \\
    \midrule
    \multirow{6}{*}{$\checkmark$} & \multirow{6}{*}{$\times$} & \multirow{6}{*}{$\times$}
    & Qwen2.5-72B & 0.707 & 0.716 & 1.000 & 1.000 \\
    &&& Llama-3.3-70B & 0.678 & 0.620 & 1.000 & 0.999 \\
    &&& Llama-3.1-70B & 0.674 & 0.672 & 1.000 & 1.000 \\
    &&& Gemma-3-27B & 0.652 & 0.627 & 1.000 & 1.000 \\
    &&& Gemma-3-12B & 0.637 & 0.585 & 1.000 & 1.000 \\
    &&& Qwen2.5-32B & 0.622 & 0.645 & 0.999 & 1.000 \\
    \midrule
    \multirow{2}{*}{$\checkmark$} & \multirow{2}{*}{$\times$} & \multirow{2}{*}{$\checkmark$}
    & GPT-OSS-120B & 0.632 & 0.617 & 0.998 & 1.000 \\
    &&& GPT-OSS-20B & 0.601 & 0.590 & 1.000 & 0.994 \\
    \midrule
    \multirow{5}{*}{$\checkmark$} & \multirow{5}{*}{$\checkmark$} & \multirow{5}{*}{$\times$} 
    & Swallow-3.1-70B & 0.698 & 0.698 & 1.000 & 1.000 \\
    &&& Swallow-3.3-70B & 0.692 & 0.731 & 1.000 & 1.000 \\
    &&& ABEJA-V2-32B & 0.687 & 0.694 & 0.992 & 1.000 \\
    &&& LLM-jp-3.1-13B & 0.582 & 0.244 & 1.000 & 0.202 \\
    &&& LLM-jp-3.1-8x13B & 0.495 & 0.320 & 1.000 & 0.836 \\
    \bottomrule
    \end{tabular}
    \caption{F1 scores and Faithfulness scores across models, zero- and four-shot settings, and dataset splits. O indicates whether the model is open-weight, J indicates whether the model is Japanese-specialised, and R indicates whether the model is a reasoning model. The models are sorted by their F1 scores in the zero-shot setting within groups sharing the same O, J, and R values.}
    \label{tab:results}
\end{table*}
\begin{table*}[ht!]
    \centering
    \small
    \begin{tabular}{llllcccccc}
    \toprule
    \multirow{2}{*}{O} & \multirow{2}{*}{J} & \multirow{2}{*}{R} & \multirow{2}{*}{Model} & \multicolumn{3}{c}{Zero-shot} & \multicolumn{3}{c}{Four-shot} \\
    \cmidrule(rl){5-7}\cmidrule(rl){8-10}
    &  &  &  & Cvl & Cst & Pen & Cvl & Cst & Pen \\
    \midrule
    \multirow{7}{*}{$\times$} & \multirow{7}{*}{$\times$} & \multirow{7}{*}{$\times$} 
      & GPT-4.1 & 0.722 & 0.739 & 0.719 & 0.729 & 0.758 & 0.715 \\
    &&& GPT-4o-11 & 0.719 & 0.720 & 0.668 & 0.740 & 0.750 & 0.687 \\
    &&& Haiku 3.5 & 0.060 & 0.238 & 0.282 & 0.645 & 0.724 & 0.675 \\
    &&& Opus 3 & 0.664 & 0.783 & 0.656 & 0.698 & 0.790 & 0.658 \\
    &&& Opus 4.1 & 0.617 & 0.733 & 0.405 & 0.795 & 0.876 & 0.741 \\
    &&& Sonnet 4 & 0.668 & 0.826 & 0.636 & 0.712 & 0.861 & 0.692 \\
    &&& Sonnet 4.5 & 0.028 & 0.184 & 0.011 & 0.733 & 0.872 & 0.685 \\
    \midrule
    \multirow{6}{*}{$\times$} & \multirow{6}{*}{$\times$} & \multirow{6}{*}{$\checkmark$}
      & GPT-5 & 0.779 & 0.845 & 0.791 & 0.767 & 0.846 & 0.771 \\
    &&& Opus 4.1 (w/ Reasoning) & 0.816 & 0.905 & 0.715 & 0.853 & 0.918 & 0.831 \\
    &&& Sonnet 4 (w/ Reasoning) & 0.757 & 0.847 & 0.784 & 0.762 & 0.835 & 0.766 \\
    &&& Sonnet 4.5 (w/ Reasoning) & 0.016 & 0.284 & 0.045 & 0.821 & 0.889 & 0.796 \\
    &&& o3 & 0.757 & 0.831 & 0.746 & 0.731 & 0.843 & 0.776 \\
    &&& o4-mini & 0.661 & 0.718 & 0.663 & 0.648 & 0.740 & 0.686 \\
    \midrule
    \multirow{6}{*}{$\checkmark$} & \multirow{6}{*}{$\times$} & \multirow{6}{*}{$\times$}
      & Gemma-3-12B & 0.632 & 0.676 & 0.613 & 0.561 & 0.675 & 0.560 \\
    &&& Gemma-3-27B & 0.632 & 0.712 & 0.647 & 0.599 & 0.698 & 0.633 \\
    &&& Llama-3.1-70B & 0.680 & 0.680 & 0.651 & 0.667 & 0.710 & 0.651 \\
    &&& Llama-3.3-70B & 0.678 & 0.697 & 0.660 & 0.565 & 0.735 & 0.650 \\
    &&& Qwen2.5-32B & 0.581 & 0.710 & 0.650 & 0.608 & 0.727 & 0.667 \\
    &&& Qwen2.5-72B & 0.695 & 0.746 & 0.707 & 0.707 & 0.750 & 0.710 \\
    \midrule
    \multirow{2}{*}{$\checkmark$} & \multirow{2}{*}{$\times$} & \multirow{2}{*}{$\checkmark$}
      & GPT-OSS-120B & 0.612 & 0.701 & 0.627 & 0.611 & 0.666 & 0.594 \\
    &&& GPT-OSS-20B & 0.590 & 0.655 & 0.583 & 0.569 & 0.636 & 0.606 \\
    \midrule
    \multirow{5}{*}{$\checkmark$} & \multirow{5}{*}{$\checkmark$} & \multirow{5}{*}{$\times$} 
      & ABEJA-V2-32B & 0.671 & 0.748 & 0.673 & 0.673 & 0.773 & 0.682 \\
    &&& LLM-jp-3.1-13B & 0.577 & 0.619 & 0.563 & 0.096 & 0.471 & 0.343 \\
    &&& LLM-jp-3.1-8x13B & 0.452 & 0.595 & 0.513 & 0.254 & 0.512 & 0.307 \\
    &&& Swallow-3.1-70B & 0.700 & 0.711 & 0.681 & 0.677 & 0.751 & 0.710 \\
    &&& Swallow-3.3-70B & 0.699 & 0.698 & 0.666 & 0.725 & 0.772 & 0.711 \\
    \bottomrule
    \end{tabular}
    \caption{F1 scores on every subject across models, zero- and four-shot settings and dataset splits. O, J, and R share the meaning with tables above. Cvl, Cst, and Pen represent the Civil Code, the Constitution, and the Penal Code, respectively.}
    \label{tab:subject_wise_f1}
\end{table*}
Table~\ref {tab:results} presents the F1 and Faithfulness scores on the whole dataset, under zero- and four-shot settings, grouped by whether a model is open~(O), is Japanese-specialised~(J), and supports reasoning~(R).

Overall, proprietary models show better performance than open-weight models on the JBE-QA dataset.
Proprietary models achieve F1 scores ranging from 0.602 to 0.861, except outliers Haiku-3.5 and Sonnet-4.5.
Compared with the proprietary models, open-weight models show lower scores, with F1 scores ranging from 0.320 to 0.731.

The low F1 scores of Haiku-3.5 and Sonnet-4.5 stem from their low faithfulness scores.
Most models are able to follow the instruction, given the majority of faithfulness scores exceed 0.990.
However, Haiku-3.5 and Sonnet-4.5 have a faithfulness score of 0.259 and 0.027, respectively.
A common pattern in the outputs of Haiku-3.5 and Sonnet-4.5 is that they provide reasons after the prediction, even though we instructed them to output only a binary value. 
For example, ``1. Explanation: According to the position of the case law, this case constitutes the crime of evidence destruction.''.
When a model's output is not a binary value as instructed, its prediction is set to 0 by default~(c.f. Section~\ref{sec:baseline_experiment}).
This results in an increase of false negatives and, consequently, a decrease in F1 score.

For proprietary models, providing exemplars generally improves their performance.
This trend is particularly noticeable for Haiku-3.5 and Sonnet-4.5, whose F1 scores increase by more than 0.500 in the four-shot setting.
In contrast, eight out of thirteen open-weight models do not show higher scores in the four-shot setting than in the zero-shot setting.
For example, Llama-3.3 has an F1 score of 0.620 in the four-shot setting, which is 0.058 lower than in the zero-shot setting.

For LLM-jp-3.1 models, exposure to exemplars even substantially decreases their performance.
The F1 score of LLM-jp-3.1-13B drops from around 0.582 in the zero-shot setting to around 0.244 in the four-shot setting.
For LLM-jp-3.1-8x13B, it decreases from 0.495 to 0.320.
This decline is attributed to the models' tendency to generate content that is similar to the provided exemplars.
An inspection of the LLM-jp 13B outputs shows that it tends to answer the input question in binary value first.
Subsequently, it generates content regarding cash and extortion, which exactly matches the topic of the last exemplar.
Consequently, many of its predictions are set to 0, leading to a large drop in faithfulness scores and hence a lower F1 score.

In short, while exemplars might provide clues for proprietary models, they could instead distract open-weight models from solving the task.
This difference may stem from the high-quality and large-scale corpora used to pretrain and fine-tune proprietary models.
Such data exposure likely helps these models better capture the semantic content of exemplars and follow the intended format, rather than being misled by their superficial information.

Proprietary reasoning models outperform their non-reasoning counterparts.
For reasoning models, the F1 scores range from 0.077 to 0.814 in the zero-shot setting and from 0.673 to 0.861 in the four-shot setting.
For non-reasoning models, the performance ranges from 0.055 to 0.724 and from 0.666 to 0.799 under zero-shot and four-shot settings, respectively.
However, for open-weight models, the performance gain brought by reasoning is not as strong as for proprietary models.
This can be observed from the fact that the GPT-OSS models do not always outperform the open non-reasoning models.

The Japanese specialisation leads to a marginal improvement under the zero-shot setting.
Swallow-3.1 and Swallow-3.3, which are derived from Llama-3.1 and 3.3, respectively, via continued pretraining on Japanese corpora, show an advantage of less than 0.03.
ABEJA-V2 outperforms its base model, Qwen2.5-32B, by 0.065.
The advantage of Japanese specialisation becomes most evident for Llama-3.3 under the four-shot setting.
Swallow-3.3 outperforms Llama-3.3 by 0.111, but for other models, the improvement is less than 0.050.

The performance of the models trained on Japanese corpora from scratch is unsatisfactory.
The F1 scores of LLM-jp-3.1 models remain lower, staying less than 0.600 in the zero-shot setting and less than 0.350 under the four-shot setting.
These scores are close to, or even below, the F1 score expected from random guessing, which is 0.488, given our task is a binary classification with 47.6\% positive instances.

\subsection{Performance on Subjects}
Table~\ref{tab:subject_wise_f1} shows the F1 scores for each subject.
In both zero-shot and four-shot settings, the performance on the Constitution is always the highest or the second highest.
For example, Opus-4.1 with reasoning reaches an F1 score above 0.900 for the Constitution in both four-shot and zero-shot settings.
All models evaluated have relatively good knowledge of the Constitution.

By contrast, across models, the Civil Code and the Penal Code are consistently the subjects that models struggle with, regardless of model family or zero- and four-shot settings. 
For the zero-shot setting, 11 out of 26 models show the poorest performance on the Civil Code, whereas 15 on the Penal Code.
This pattern does not change largely for the four-shot setting: 14 models show the poorest performance on the Civil Code, and 12 on the Penal Code.

Comparing the statistics under zero- and four-shot settings, the subject where models perform the worst changes only marginally, regardless of whether exemplars are provided.
In contrast, the reasoning capability of models contributed to their performance.
Among seven proprietary non-reasoning models, six perform worst on the Penal Code, whereas among six proprietary reasoning models, four perform the worst on the Civil Code.
Reasoning generally improves model performance, but it may benefit different subjects to varying degrees.

\subsection{Case Study}
Among all instances, there is a particularly difficult one that none of the models is able to solve, regardless of whether it is under a zero-shot or few-shot setting.
We present the instance in English translation below.
\begin{quote}
    Is the following statement concerning the Crimes of Arson and Fire Caused through Negligence correct or not?\\
    A person \textit{A} lived alone in a house that he owned. The house was covered by fire insurance. Intending to obtain the insurance payout on the house fraudulently, he set fire to his own house, completely burning it down and thereby endangering the public. In this case, \textit{A} is guilty of the crime of Arson of Uninhabited Buildings belonging to another person.
\end{quote}

The statement in the question is about the crime of ``Arson of Uninhabited Buildings belonging to another person'' (Art.\/~109 of the Penal Code).
The correct answer to the question is True according to both the interpretation of the precedents and the related articles.

First, the phrase "Uninhabited Buildings" in this case is not used in its literal sense.
Instead, its interpretation has been established in the precedents.
Specifically, ``Uninhabited Buildings'' refers to buildings that are not being inhabited by anyone other than the offender.
In this case, because the inhabitant \textit{A} is the very person who set the building on fire, this building is considered ``Uninhabited''. 

Second, Art.\/~115 of the Penal Code defines a special case regarding the ``Arson of Uninhabited Buildings belonging to another person''. 
If the owner sets fire to their own building and the building is insured, the owner is considered to be an offender of another person's building. 
Thus, in this case \textit{A} is considered to set fire to ``buildings belonging to another person''.

If a model lacks precise knowledge of related articles in the Penal Code and their judicial interpretation established in the precedents and instead relies only on the superficial information of the phrases ``Uninhabited'' and ``belonging to another person'', it might produce an incorrect judgment.

We inspect the reasoning part of the best-performing model Opus-4~(w/ Reasoning).
In its reasoning outputs, Opus-4 raises the following reasons:  
``\textit{A}'s house was his own property, not that of another person'', 
and ``Since \textit{A} lived there alone, it was an inhabited building, not an uninhabited one'',
Similar reasoning patterns can be observed in other models as well.
Therefore, the evaluated models are influenced by superficial information and might not properly possess the knowledge of the related article and precedents.

\section{Related Work}
There are different benchmarks, from focusing on single tasks in specific jurisdictions to employing multiple tasks, languages, and legal systems, reflecting the growing needs of computational solutions for the legal domain.

For English resources, LexGLUE~\cite{chalkidis2022lexglue}, provided a suite of tasks for legal language understanding. 
This was followed by LegalBench~\cite{guha2023legalbench}, a collaboratively built benchmark with 162 tasks focused on legal reasoning across the Constitution, the Penal Code, and the Civil Code of the United States.
Moreover, there are datasets focused on specific tasks like legal judgment prediction, with datasets such as PILOT~\cite{cao2024pilot} and ClassActionPrediction~\cite{semo2022classactionprediction}.

In Europe, MultiEURLEX~\cite{chalkidis2021multieurlex} provided a multilingual corpus of EU laws for document topic classification, covering 23 languages.
LEXTREME~\cite{niklaus2023lextreme} expanded on this with 11 datasets covering 24 languages, providing a multi-task benchmark for the legal domain.
Other notable European resources included EUR-Lex-Sum for legal document summarisation ~\cite{aumiller-etal-2022-eur} and benchmarks based on Swiss legal document~\cite{rasiah2024one}.

Asian legal systems have also seen a surge in benchmark development.
For Chinese, JEC-QA~\cite{zhong2020jec} is a large-scale question-answering dataset from the National Judicial Examination. 
More recent and comprehensive benchmarks include LawBench~\cite{fei2024lawbench} with 20 tasks, LexEval~\cite{li2024lexeval} with 23 tasks, and LAiW~\cite{dai2025laiw}.
For Indian law, IL-TUR~\cite{joshi2024iltur} provided a multilingual benchmark, and Nyayaanumana~\cite{nigam2025nyayaanumana} is a large judgment prediction dataset.
In Korea, LBOX OPEN~\cite{hwang2022multi} and KBL~\cite{kimyeeun2024developing} are prominent multi-task benchmarks.
Southeast Asian languages are also represented, with benchmarks for Thai~\cite[NitiBench,][]{akarajaradwong2025nitibench}, Vietnamese~\cite[VLQA,][]{nguyen2025vlqa}, and Indonesian~\cite[IndoLER,][]{yulianti2024named}.

In Japan, there are a few benchmark datasets available for the legal domain.
COLIEE~\cite{goebel2024overview} employed the subset of questions from the \textit{tantō-shiki} test.
They utilised the Civil Code questions and designed their task as a textual entailment task and information retrieval.
\citet{2023GPT-barexam} also employed the \textit{tantō-shiki} test to evaluate the performance of GPT-3~\cite{10.5555/3495724.3495883}, GPT-3.5-turbo, and GPT-4~\cite{openai2023gpt4} in the Japanese legal domain. 
They tasked the models to solve the multiple-choice questions as-is in the \textit{tantō-shiki} test. 
They also provided binary questions converted from the \textit{tantō-shiki} similar to our dataset, but only available for the Civil Code.
More recently, the Japanese Tort-case Dataset (JTD)~\cite{yamada2024japanese} provided a dataset for Japanese Legal Judgment Prediction (LJP), focusing on tort cases from the Civil Code of Japan. 
However, these datasets are limited in scope, primarily focusing on the Civil Code. 
Our work introduces the first comprehensive benchmark that covers not only the Civil Code but also the Penal Code and the Constitution, addressing a significant gap in resources for evaluating LLMs on Japanese legal texts.

\section{Conclusion}
We introduce JBE-QA, a dataset for evaluating LLMs' legal knowledge. 
It covers the Civil Code, the Penal Code, and the Constitution, extending beyond the Civil Code focus of the previous Japanese legal resources, and hence enabling a comprehensive legal knowledge evaluation. 
We benchmark 26 models, including proprietary, open-weight, Japanese-specialised, and reasoning models, under zero-shot and four-shot settings.
Hence, we establish the baseline of legal knowledge quality of the current models.
The results show that reasoning and proprietary models perform best, Japanese specialisation improves few-shot performance, and the Constitution is easier than the Civil Code or the Penal Code, highlighting the possible directions of future improvements.

\section{Ethics Statement}
We believe that this dataset does not contain any significant ethical concerns. 
Our source of the dataset is an official archive of the past Japanese bar exam, and it does not contain any private information or confidential information. 
Currently, there are no feasible use cases for this dataset that can involve potential harm or ethical issues.

\section{Limitations}
Our dataset primarily focuses on the fundamental legal knowledge of the Japanese legal system using \textit{tantō-shiki} questions.  
The \textit{tantō-shiki} questions require relatively fewer reasoning steps in order to apply the knowledge to solve the questions, compared with \textit{ronbun-shiki} questions. 
In the practical situation, more intensive legal reasoning is required to apply the legal knowledge correctly to a longer context. 
To evaluate such higher-level reasoning and knowledge application, \textit{ronbun-shiki} questions would be a good device.
The difficulty of evaluating the \textit{ronbun-shiki} questions, even with human experts, restricts the direct implementation of them as a benchmark task; however, it would be an interesting future direction.

Another limitation of the current version of this dataset is that the questions are not automatically updated even if the relevant laws are revised. 
Thus, there is a risk that the gold labels in the dataset might not align with future versions of laws. 
Developing an automatic update system for questions affected by revised laws would be an ideal solution.

\section{Acknowledgement}
This work was supported by the ``R\&D Hub Aimed at Ensuring Transparency and Reliability of Generative AI Models'' project of the Ministry of Education, Culture, Sports, Science and Technology, and by JST PRESTO, Japan, Grant Number JPMJPR236B.

\section{Bibliographical References}\label{sec:reference}
\bibliographystyle{lrec2026-natbib}
\bibliography{lrec2026-example,legal_benchmark}

\end{document}